\def\year{2021}\relax
\documentclass[letterpaper]{article} 
\usepackage{aaai21}  
\usepackage{times}  
\usepackage{helvet} 
\usepackage{courier}  
\usepackage[hyphens]{url}  
\usepackage{graphicx} 
\usepackage{natbib}  
\usepackage{caption} 
\usepackage{amsmath} 
\usepackage{amssymb} 
\usepackage{bbm} 
\usepackage{algorithm} 
\usepackage{algpseudocode}

\usepackage{amsthm} 
\usepackage{color}

\newtheorem{theorem}{Theorem}[section]
\newtheorem{lemma}[theorem]{Lemma}
\newtheorem{corollary}{Corollary}[theorem]

\newif\ifflow\flowtrue
\newif\ifdebugdoc\debugdocfalse

\ifdebugdoc
    \newcommand{\outline}[1]{\textbf{\colorbox{yellow}{Outline:}\textcolor{red}{#1.}}}

    \newcommand{\del}[1]{\textcolor{blue}{\sout{#1}}}
    \newcommand{\todo}[1]{\textcolor{red}{{\bf TODO}: {#1}}}
    
    \newcommand{\yuanjie}[1]{\textcolor{red}{Yuanjie: {#1}}}
    \newcommand{\fixme}[1]{\textcolor{blue}{{\bf FIXME}: {#1}}}
\else
    \newcommand{\outline}[1]{}

    \newcommand{\del}[1]{}
    \newcommand{\todo}[1]{}
    
    \newcommand{\yuanjie}[1]{}
    \newcommand{\fixme}[1]{}

\fi

\ifflow
    \newcommand{\p}[1]{\vskip 1ex\noindent\colorbox{yellow}{\parbox{\columnwidth}{\textbf{Point:} {#1}}}}
    \newcommand{\key}[1]{\vskip 1ex\noindent\colorbox{yellow}{\parbox{\columnwidth}{\textbf{Keywords:} {#1}}}}
    \newcommand{\q}[1]{\vskip 1ex\noindent\colorbox{cyan}{\parbox{\columnwidth}{\textbf{Question:} {#1}}}}
\else
    \newcommand{\p}[1]{}
    \newcommand{\key}[1]{}
    \newcommand{\q}[1]{}
\fi

\def\name{BaTT\xspace}
\def\ie{i.e.\xspace}
\def\eg{e.g.\xspace}

\def\etc{etc\xspace}
\def\wrt{w.r.t.\xspace}

\usepackage{url}
\usepackage{subfig}
\usepackage{xcolor}
\usepackage{xspace}
\usepackage{amssymb}
\usepackage{amsmath}
\usepackage{balance}
\usepackage{booktabs}
\usepackage{tabularx}
\usepackage{multirow}
\usepackage{enumitem}
\usepackage{multirow}
\usepackage{graphicx}
\usepackage{amsfonts}
\usepackage[export]{adjustbox}
\usepackage{algorithm}
\usepackage{wrapfig}
\usepackage{algpseudocode}
\makeatletter
\renewcommand{\ALG@beginalgorithmic}{\scriptsize}
\makeatother
\usepackage[normalem]{ulem}
\usepackage{varwidth}
\usepackage{amsmath}
\DeclareMathOperator*{\argmax}{arg\,max}
\DeclareMathOperator*{\argmin}{arg\,min}

\algnewcommand\algorithmicswitch{\textbf{switch}}
\algnewcommand\algorithmiccase{\textbf{case}}
\algnewcommand\algorithmicassert{\texttt{assert}}
\algnewcommand\Assert[1]{\State \algorithmicassert(#1)}%
\algnewcommand{\IIf}[1]{\State\algorithmicif\ #1\ \algorithmicthen}
\algnewcommand{\IElse}[1]{\State\algorithmicelse\ #1\ \algorithmicthen\ }
\algnewcommand{\EndIIf}{\unskip\ \algorithmicend\ \algorithmicif}
\algdef{SE}[SWITCH]{Switch}{EndSwitch}[1]{\algorithmicswitch\ #1\ \algorithmicdo}{\algorithmicend\ \algorithmicswitch}%
\algdef{SE}[CASE]{Case}{EndCase}[1]{\algorithmiccase\ #1}{\algorithmicend\ \algorithmiccase}%
\algtext*{EndSwitch}%
\algtext*{EndCase}%

\title{Bandit Policies for Reliable Cellular Network Handovers in Extreme Mobility}

\author{

     Yuanjie Li \textsuperscript{\rm 1},
     Esha Datta \textsuperscript{\rm 2},
     Jiaxin Ding \textsuperscript{\rm 3},
     Ness Shroff \textsuperscript{\rm 4},
     Xin Liu \textsuperscript{\rm 2}
    \\
}
\affiliations{

    \textsuperscript{\rm 1}Tsinghua University,
    \textsuperscript{\rm 2}University of California, Davis, 
     \textsuperscript{\rm 3}Shanghai Jiao Tong University, 
    \textsuperscript{\rm 4}Ohio State University\\

    liyuanjie08@gmail.com,
    edatta@ucdavis.edu,
    jiaxinding@sjtu.edu.cn, shroff.11@osu.edu, xinliu@ucdavis.edu

}

\newcommand{\xl}[1]{\textcolor{red}{LX:#1}}

\newcommand{\ed}[1]{\textcolor{blue}{ED:#1}}

\begin{document}
\maketitle

\begin{abstract}

The demand for seamless Internet access under extreme user mobility, such as on high-speed trains and vehicles, has become a norm rather than an exception. However, the 4G/5G mobile network is not always reliable to meet this demand, with non-negligible failures during the handover between base stations. A fundamental challenge of reliability is to balance the {\em exploration} of more measurements for satisfactory handover, and {\em exploitation} for timely handover (before the fast-moving user leaves the serving base station's radio coverage). This paper formulates this trade-off in extreme mobility as a composition of two distinct multi-armed bandit problems. We propose Bandit and Threshold Tuning (\name) to minimize the regret of handover failures in extreme mobility. 
\name uses $\epsilon$-binary-search to optimize the threshold of the serving cell's signal strength to initiate the handover procedure with $\mathcal{O}(\log J \log T)$ regret.
It further devises opportunistic Thompson sampling, which optimizes the sequence of the target cells to measure for reliable handover with $\mathcal{O}(\log T)$ regret.
Our experiment over a real LTE dataset from Chinese high-speed rails validates significant regret reduction and a 29.1\% handover failure reduction.

\end{abstract}

\section{Introduction}
\label{sec:introduction}

The widespread adoption of high-speed rail has made extreme mobility increasingly common. Today, high-speed trains can move up to 350 km/hr and passengers require always-on Internet access. But with great speed comes great handover failures: empirical studies of 4G LTE from high-speed rail shows that the network failure ratio can range 5.5\% to 12.6\% (Table~\ref{tab:failure-ratio}), which is about 2$\times$ higher than low-mobility scenarios such as walking and driving \cite{li2020beyond}. Clearly, while the existing mobile network can successfully support billions of stationary or low-mobility users, it struggles to maintain the same level of reliable service for users with extreme mobility. 

This is because radio base stations face a unique and fundamental challenge in handovers under extreme mobility: how to balance the need to take more measurements (exploration) with the decision to make a timely, successful handover for a fast-moving user (exploitation). Motivated by this challenge, we formulate the reliable handover in extreme mobility as a composition of two distinct multi-armed bandit problems. To the best of our knowledge, this is the first time the reliability problem has been formulated in this manner. We then present a novel solution, Bandit and Threshold Tuning
(\name). This ifrst identifies an optimal serving cell threshold value and, under this condition, determines, for each mobile user, {\em when} and in {\em what sequence} to take a measurement, and when to execute a handover.

\name has two routines. To determine {\em when} to start the measurements for handover, we must identify the optimal serving cell threshold value that balances the exploration-exploitation dilemma in extreme mobility. We formulate it as a closest sufficient arm identification problem. Specifically, we define this as a $J$-armed stochastic bandit problem over $T$ rounds with the following characteristics: 1) the agent is given a threshold value $R$; 2) the available arms are ordered in expected reward; and 3) the goal of the agent is to find the optimal arm (that is, the arm whose expected reward is closest to and greater than $R$). We introduce the algorithm $\epsilon$-Binary-Search-First, which is a variation of the well-established bandit algorithm $\epsilon$-search-first, to solve this problem with regret on the order of $\mathcal{O}(\log J \log T)$. 

Then, to optimize the handover target with high reliability, \name decides {\em what sequence} of target cells to measure.
This can be formulated as an opportunistic bandit with side observations.
For each serving cell, we define a stochastic bandit problem with $K$ arms (neighboring cells) over $N$ rounds (measurements) where 1) the best arm is fixed over every round; 2) the exploration cost of a suboptimal arm depends on a round-variant external condition; and 3) the round-variant external condition determines the manner in which the next arm to be pulled is chosen. 
We introduce the opportunistic Thompson sampling algorithm to solve this problem with $\mathcal{O}(\log T)$ regret. 
We evaluate \name with a large-scale LTE dataset on the Chinse high-speed trains. Our result shows $\epsilon$-Binary-Search-First outperforms a uniform search of threshold values. Further, we achieve significantly lower regret than UCB and Thompson sampling and reduce 29.1\% handover failures compared to the state-of-the-art 4G/5G handover policies. 

\section{4G/5G Mobility Management Policy Today}
\label{sec:back}

The 4G LTE and 5G cellular networks are the largest wireless infrastructure that, together with wired Internet, enable ubiquitous Internet access and {\em wide-area} mobility management for users. 4G/5G deploys base stations (``cells'') in different geographical areas. 
When a user leaves one base station's radio coverage, it is migrated to another base station (a {\em handover}) to retain its network service.

\begin{figure}[t]
\centering
\includegraphics[width=.8\columnwidth,valign=c]{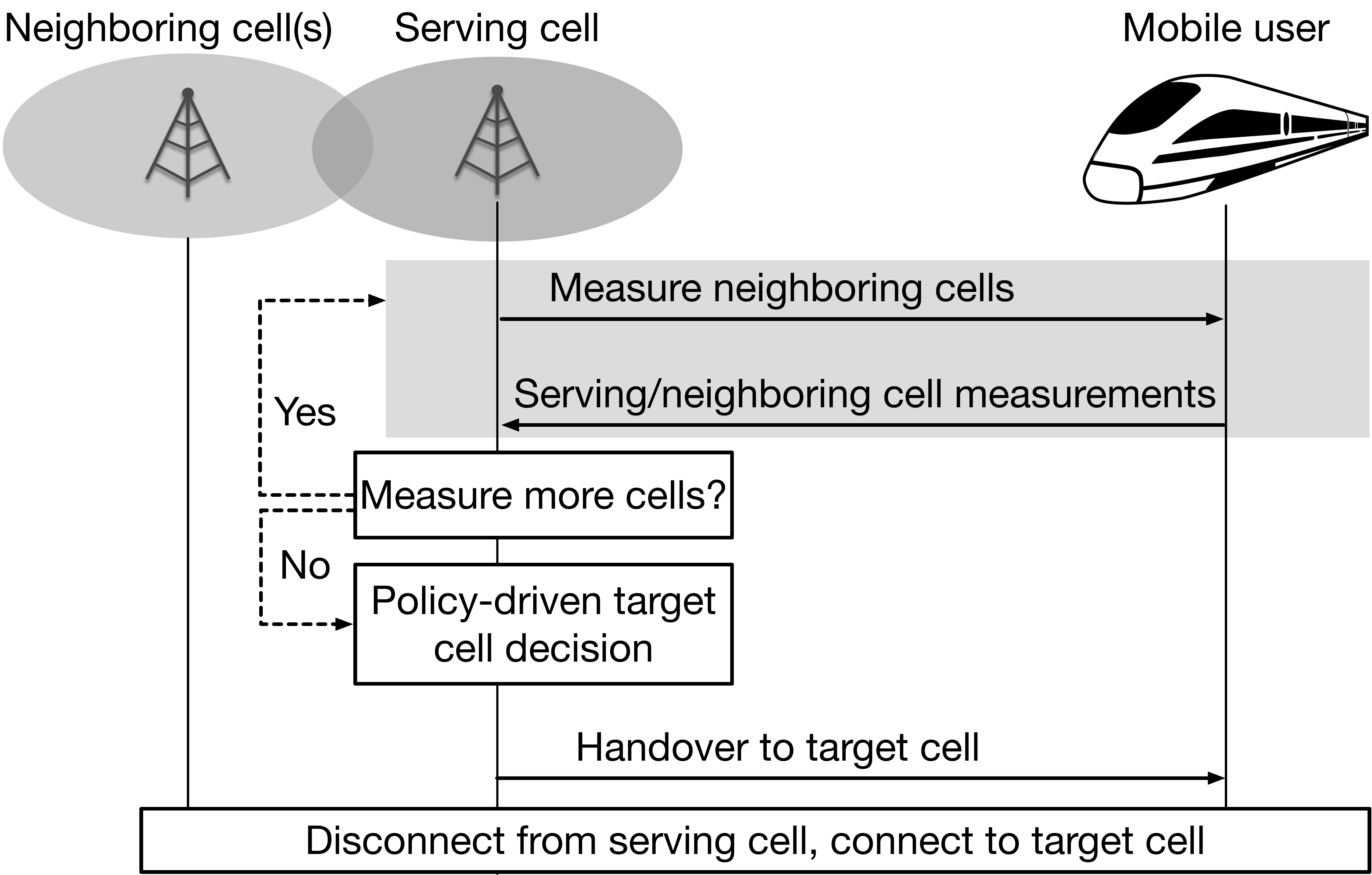}
\caption{Extreme mobility management in 4G/5G today.}
\label{fig:back}
\label{fig:back:procedure}
\end{figure}

Figure~\ref{fig:back:procedure} shows the standard 4G/5G handover procedure \cite{TS36.331, TS38.331}. 
When a mobile user connects to a serving cell, it receives a list of neighboring cells.
The user can then measure these neighboring cells' signal strengths one by one. If any neighboring cell satisfies the standard triggering criteria (\eg, a neighboring cell's signal strength is offset better than the serving cell's \cite{TS36.331, TS38.331}), the user will report this cell's and serving cell's signal strengths to the serving cell. 
The serving cell will then run its local policy to decide if more neighboring cells should be measured, whether handover should begin, and which target cell the user should hand over to. 
If the serving cell chooses to take new measurements, it will provide the user with a new neighboring cell list. If it chooses to handover, the serving cell will (in coordination with the target cell)  send the handover command with the target cell's identifier to the user. The user will disconnect from the serving cell and connect to the target cell.

\begin{table}[t]
\centering
\resizebox{\columnwidth}{!}{
		\begin{tabular}{||l|l|l|l||}
		\hline
		{\bf User speed (km/h)} & 200 & 300 & 350\\
		\hline
		{\bf Total handover failures} & 5.5\% {\tiny (100\%)} & 12.1\% {\tiny (100\%)} & 12.6\% {\tiny (100\%)}\\
		{\bf \ \ \  $\bullet$ Due to serving cell} & 4.9\% {\tiny (90.0\%)}& 9.3\% {\tiny (77.1\%)} & 11.0\% {\tiny  (87.3\%)}\\
		{\bf \ \ \ $\bullet$ Due to target cell} & 0.6\% {\tiny (10.0\%)} & 2.8\% {\tiny (22.9\%)} & 1.6\% {\tiny (12.7\%)}\\
		\hline
		\end{tabular}}
\vspace{-3mm}
\caption{Handover failures in extreme mobility}
\label{tab:failure-ratio}
\end{table}

\subsubsection{Is 4G/5G handover reliable in extreme mobility?}
The current 4G/5G handover design is primarily meant for static and low-mobility scenarios. Recent studies \cite{li2020beyond,wang2019active} have shown that when users move at fast speeds, they experience non-negligible handover failures, thus frequently losing Internet access. Table~\ref{tab:failure-ratio} shows the 4G LTE handover failure ratios of a smartphone on a Chinese high-speed train from Beijing to Shanghai based on the dataset from ~\cite{wang2019active} (elaborated in $\S$\ref{sec:eval}). On average, 5.5\%, 12.1\% and 12.6\% handovers fail at the train speed of 200km/h, 300km/h and 350km/h, respectively. 
The failure ratio becomes higher with faster train speed.
Among these handover failures, 77.1\%--90.0\% of them are caused by the {\em late handover}, \ie, by the user not receiving the handover command from the serving cell by the time it leaves the serving cell's radio coverage. 
The remaining handover failures occur when the user receives the handover command from the serving cell, but fails to connect to the new target cell. In this case, the selected target cell is unreliable.

\subsubsection{Challenge: Exploration-Exploitation tradeoff}
\begin{figure*}[t]
\vspace{0mm}
\subfloat[Number of neighboring cells (arms)]{
\includegraphics[width=.3\textwidth,valign=b]{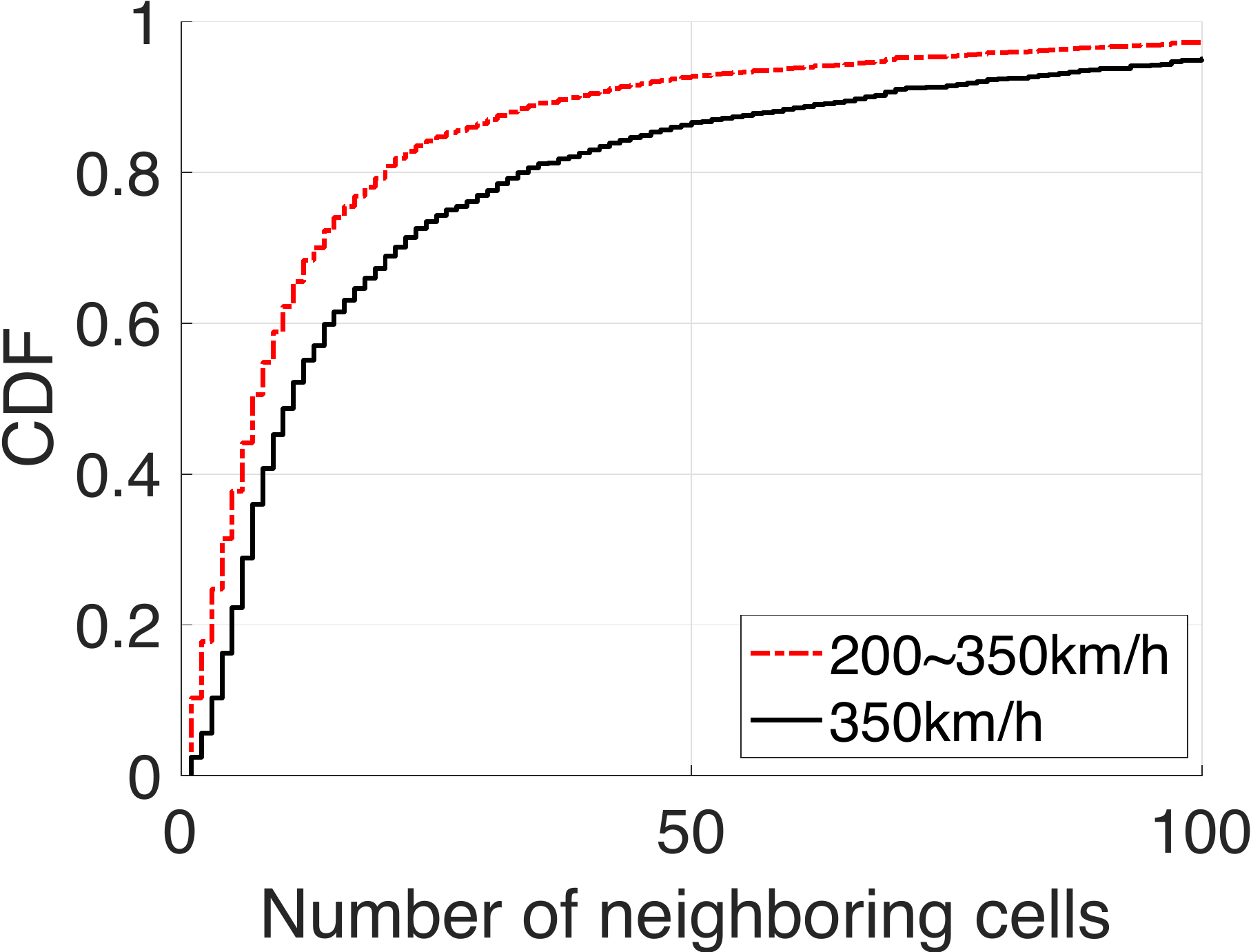}
\label{fig:cell-density}
}
\hfill
\subfloat[Failures \wrt signal strength (350km/h)]{
\includegraphics[width=.3\textwidth,valign=b]{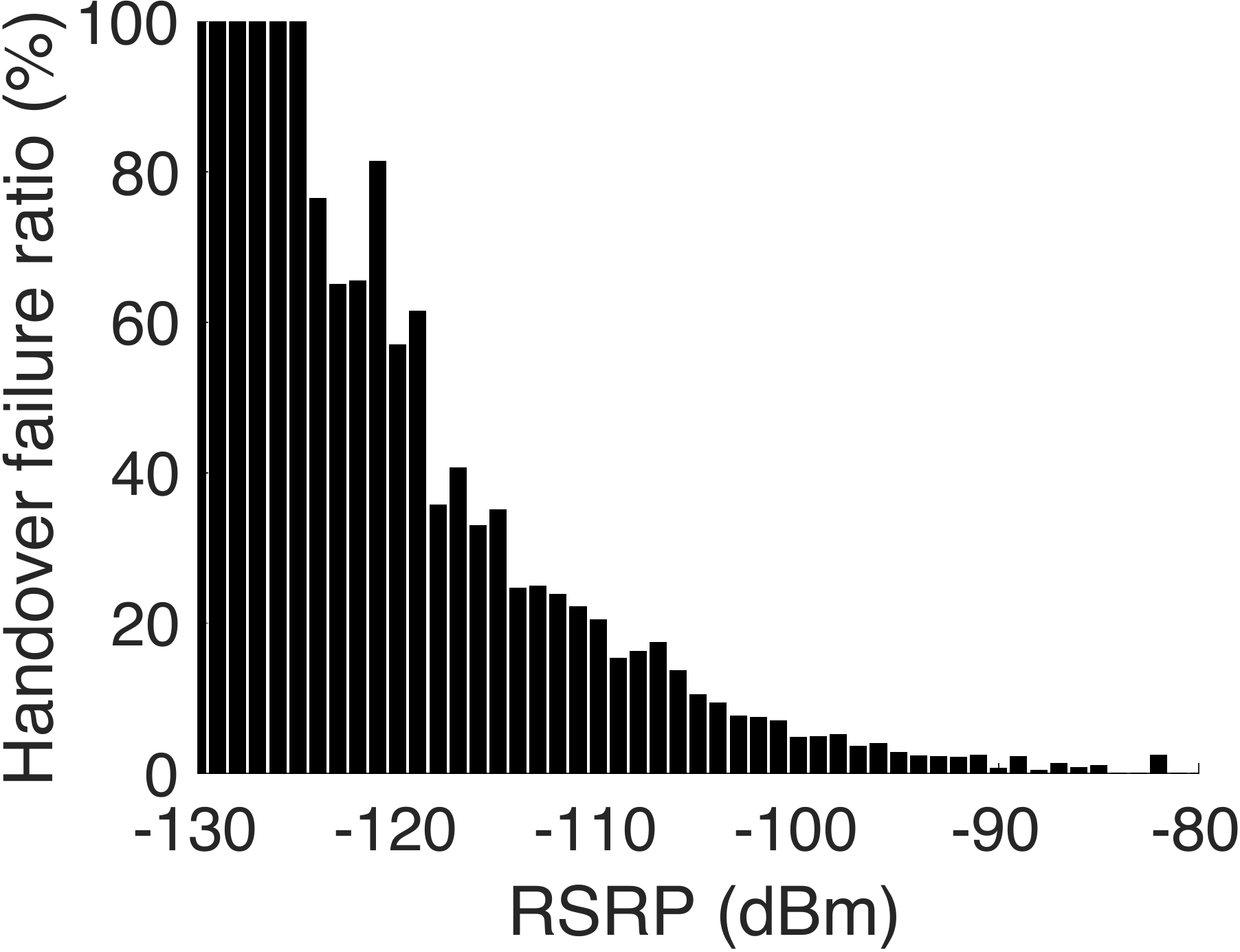}
\label{fig:failure_vs_rsrp}
}
\hfill
\subfloat[Dynamics of adjacent measurements]{
\includegraphics[width=.3\textwidth,valign=b]{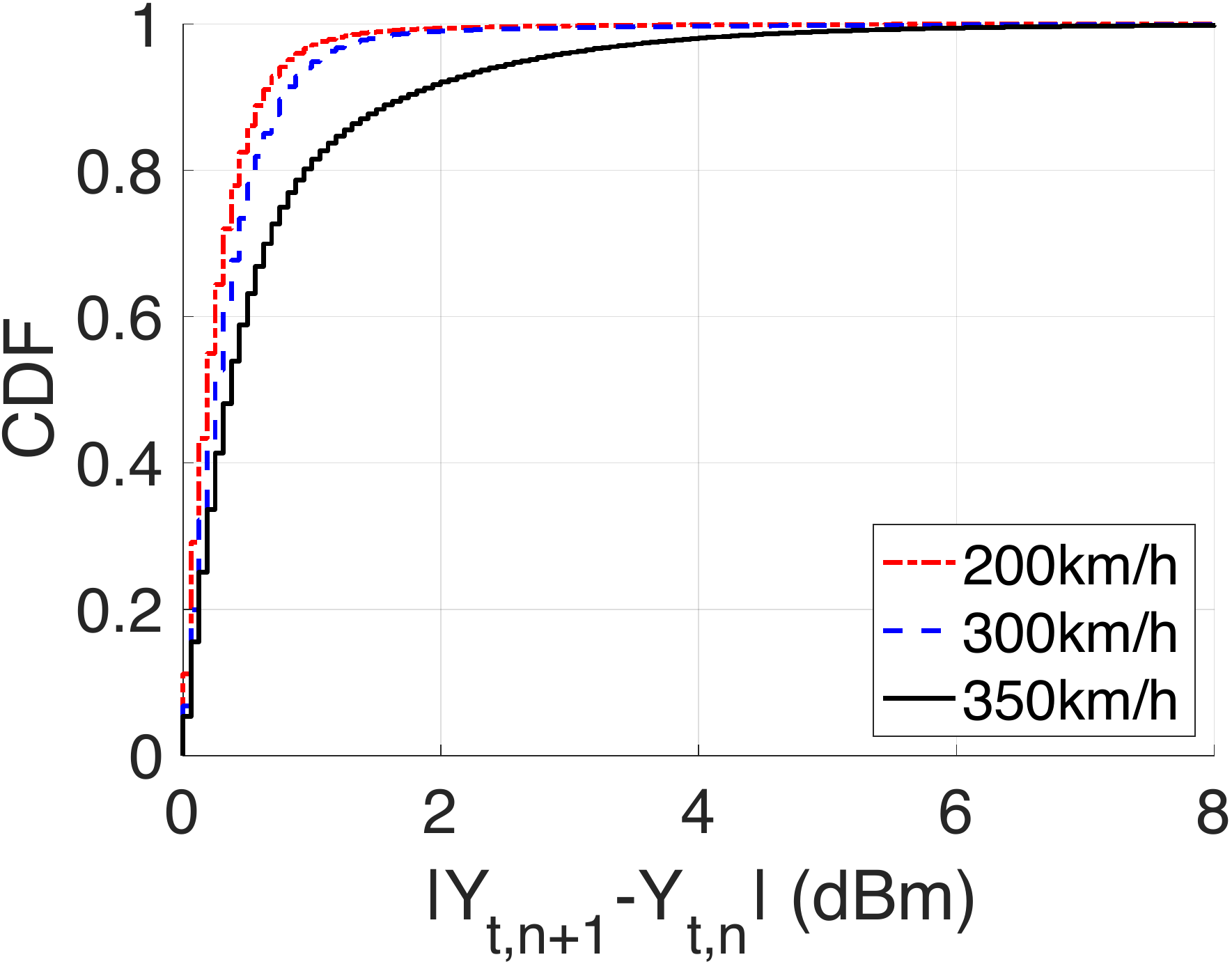}
\label{fig:rsrp_dynamics}
}
\caption{Characteristics of LTE handovers over Chinese high-speed train. 
}
\label{fig:validation}
\vspace{-5mm}
\end{figure*}

Frequent handover failures occur in extreme mobility because the serving cell faces a fundamental dilemma between {\em exploration} (more measurements for satisfactory target cell selection) and {\em exploitation} (fast measurements for timely handover).
In 4G/5G, the serving cell relies on the user to measure and report the cells' wireless signal strengths for the handover decision.  
To retain Internet access, the user must deliver
these measurements {\em before} it leaves the serving cell’s radio coverage. But, finding a reliable target cell may require scanning and measuring {\em all} available cells, in principle. Per Figure~\ref{fig:cell-density}, on average, a mobile user on a Chinese high-speed train should measure 16 different neighboring cells before making a handover decision. Note that the user has to measure these cells {\em sequentially}.  But, if the user is moving very fast, it may not be able to deliver all its measurements {\em and} trigger a handover before leaving its serving cell's radio coverage (resulting in a late handover failure).  It is reported that a user on a Chinese high-speed train moving at 350km/h takes 800ms on average to measure neighboring cells, during which it moves 78.0m along the rails \cite{li2020beyond}. This is too fast for timely handover. 
Reducing the number of cells to measure can mitigate the late handover failures. But this risks missing better target cells and therefore committing a handover to an unreliable target cell (which leads to failures).

\section{Problem Formulation}
\label{sec:problem}


As discussed in Sec.~\ref{sec:back}, 
in extreme mobility, a mobile user has a short, but critical time period to conduct effective measurements for handover. It needs to use this period efficiently by measuring the right sequence of target cells {\em before} leaving the serving cell's radio coverage. 
In other words, for reliable handover, we must answer two questions: 1) When does this critical moment start and  2) what is the right sequence of target cells to measure? 

To answer both questions, we formulate the reliable handover problem in this section. We consider a fixed serving cell with $K$ (neighboring) target cells available for  handover. We consider a sequential set of mobile users, indexed by $t$, where $t=1,\dots,T$. Our objective is to minimize the handover failure rate over all $T$ mobile users. 
Table \ref{tab:notation}  summarizes the notations we use in this paper. 


\begin{table}[t]
\centering
\resizebox{\columnwidth}{!}{
\begin{tabular}{||c | l||} 
 \hline
 Notation & Extreme Mobility Definition  \\ 
 \hline\hline
 $Z_j$ & $j$th strongest serving cell signal strength  \\ 
 \hline
 $J$ & number of available discrete serving cell signal strengths \\
 \hline
 $M$ & closest sufficient signal strength  \\
 \hline
 $R$ & predefined serving cell handover failure tolerance level  \\
 \hline
 $K$ & number of neighboring target cells \\
 \hline
 $t$ & index of mobile users \\
 \hline
 $T$ &total number of mobile users  \\
 \hline
  $n$ & index of target cell measurement for a given user\\
 \hline
 $I_{t,n}$ & $n$th target cell measured for user $t$ \\
 \hline
 $Y_{t,n}$ & serving cell signal strength  for user $t$ at the $n$th measurement \\
 \hline
 $X_{I_{t,n}}$ & target cell signal strength  for user $t$ at the $n$th measurement  \\
 \hline
 $X_{best}$ & the strongest target cell \\
 \hline
 $c$    & regularity assumption constant \\
 \hline
\end{tabular}
}
\caption{Notation Summary}
\label{tab:notation}
\vspace{-5mm}
\end{table}



\subsection{When Does the Critical Time Begin?}
\label{subsec:tuning}

We note that wireless signal strength attenuates in a quadratic manner as the distance between the transmitter and receiver increases. As a mobile user is moving away from the serving base station, its signal strength weakens. Therefore, roughly speaking, this critical time starts when the signal strength from the serving cell is at a certain threshold. 
On one hand, this threshold needs to  be high for the serving cell signal strength  to be good enough, so that  1) handover failure will not occur often due to weak serving cell signal strength; and 2) the user has sufficient time to measure target base stations and to obtain a good target cell to handover to. 
On the other hand, we want this signal strength threshold to be low 1) to avoid the so-called ``ping-pong'' effect where a user moves between two cells frequently (which incurs a lot of signaling overhead and more frequent handover failures), and 2) to avoid a false start when an actually desirable target cell is
still too far away to be measured appropriately. 


In this paper, we formulate this ``when" problem as a  \textbf{closest sufficient arm identification} problem to identify this handover threshold. Consider a given serving cell. Let  
$$\{Z_j\} = Z_1, Z_2, \ldots Z_j, \ldots, Z_{J-1}, Z_J$$
be the sequence of increasing serving cell signal strength that can be observed by a mobile user. 
There are $J$ elements in the sequence. Let $[J]$ denote the list $\{1, 2, \ldots, J\}.$

Let the random variable $f(Z_j)$ represent handover failure due to the serving cell's signal strength $Z_j$. Note $f(Z_j) \in \{0,1\}$, where 0 indicates handover failure due to the serving cell and 1 indicates a success. The probability of a handover failure due to signal strength $Z_j$ equal to $\mathbb{P}[f(Z_j) = 0] = r_j$.


Let $R$ be the predefined serving cell handover failure tolerance level. In practice, it is often set as $1\% -3\%$. 
We make a {\em monotonic handover failure ratio assumption}: we assume that as the value of $Z_j$ increases, the probability $r_j$ decreases. This is empirically validated with the high-speed rail dataset as shown in Fig.~\ref{fig:failure_vs_rsrp}.

Our {objective} is to find $M \in \{Z_j\}$ that is the smallest signal strength $Z_j$ such that $r_j \leq R.$ That is, $M$ is the lowest signal strength at which the probability of serving cell handover failure is no larger than $R$. We refer to the signal strength $M$ as the handover threshold.


\subsection{What  Sequence of Target Cells to Measure?}
\label{subsec:pur routine}
Given a handover threshold $M$, once a mobile user triggers the measurement procedure, the key issue is to decide the sequence of the target cells to measure and the time to stop measurement and start handover.



Consider a mobile user $t$. When the handover measurement procedure is triggered, the serving cell starts a sequence of measurements of neighboring cell's signal strengths, indexed by $n$.  The decision of whether to take more measurements or to execute a handover is the central exploitation-exploration dilemma faced by the serving cell. Note that the total number of measurements made by the serving cell may vary from user to user. Further, the handover terminates the sequence of measurements of target cells.



At the $n$th measurement, let $I_{t,n}$ be the index of the target cell to measure. 
The user can then observe the serving cell signal strength $Y_{t,n}$ and the target cell signal strength $X_{I_{t,n}}.$ 
Let $X_{best}$ be the strongest target cell observed for user $t$ thus far. After a sequence of $n$ measurements, if the mobile user decides to handover, it handovers to the best target cell with signal strength $X_{best}$.

Recall that $f(Y_{t,n}) \in \{0,1\}$ is the handover failure caused by serving cell signal strength $Y_{t,n}$. Similarly, we can define $g(X)$ be the handover failure caused by the target cell with signal strength $X$, where $g(\cdot)$ and $f(\cdot)$ may be distinct functions. However, in algorithm development and evaluation, we assume $f(\cdot)=g(\cdot)$ for simplicity.

The  handover failure probability of user $t$  is $\mathbb{E}[f(Y_{t,n}) g(X_{best})]$ when the handover happens after $n$ measurements and $X_{best}$ is the best target cell. 
In general, $Y_{t,n}$ decreases with $n$ as the mobile user is moving away from the serving cell. Therefore, the tradeoff is whether to make more measurements, which improves $X_{best}$, but at the risk of decreasing $Y_{t,n}$. 

The typical practice in today's cellular operation is to measure target cells following a fixed sequence and trigger actual handover when $X_{best}$ is greater than or equal to $Y_{t,n}$ plus an offset quantity determined by the network provider. 

The \textbf{objective} of the ``what sequence" question is to decide the best order of target cell measurement and when to stop measurement and start handover. 

\begin{figure}[t]
\centering
\includegraphics[width=.8\columnwidth]{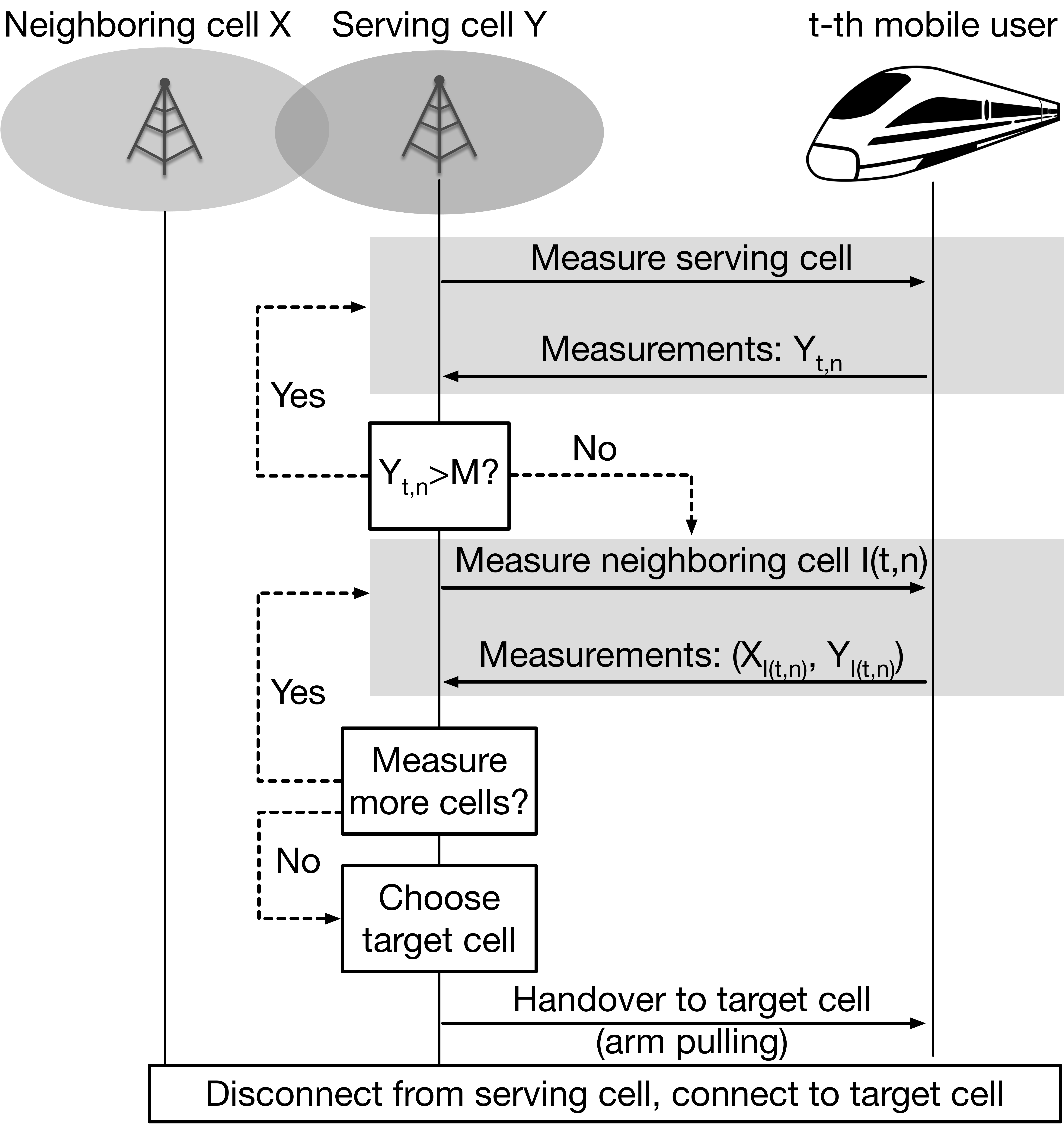}
\caption{Our bandit policy for extreme mobility.}
\label{fig:setup}
\vspace{-5mm}
\end{figure}

\section{Algorithms}
\label{sec:algorithm}


We propose \name for reliable handovers in extreme mobility. 
To solve the above two problems, BaTT combines two algorithms: $\epsilon$-Binary-Search-First and opportunistic Thompson sampling (opportunistic-TS).

\subsection{When: $\epsilon$-Binary-Search-First} 
\label{subsec:tt} 

Recall that our {objective} is to find $M$, the handover threshold, i.e., the lowest signal strength at which the probability of serving cell handover failure is no larger than $R$.   

Clearly, exploring each value of the $J$ signal strengths  is expensive. Instead, we should leverage the monotonicity property between the signal strength and handover failure rate. To do so, we propose the following  $\epsilon$-Binary-Search-First algorithm and analyze its regret in Sec.~\ref{sec:regret}. 

Specifically, $\epsilon$-Binary-Search-First is a multi-armed bandit algorithm which takes as input the number of available signal strengths (arms) $J$, 
and tuning value $R$. We provide an exploration parameter $0 \leq \epsilon \leq 1.$  Each arm $j \in [J]$ is associated with a random variable $f(Z_j)$ where 
$$\mathbb{E}[f(Z_1)] \leq \mathbb{E}[f(Z_2)] \leq \cdots \leq \mathbb{E}[f(Z_J)].$$ 
The goal of the algorithm is to identify the optimal threshold: 
\begin{equation}
M=\argmin_{Z_j} \left\{ \mathbb{E}[f(Z_j)] \geq R
\right\}
\label{eq:M}    
\end{equation}

The algorithm works in two distinct phases: exploration and exploitation. During the exploration phase, the algorithm pulls the arms in a binary search manner, which is described in the subroutine Binary-Arm-Search. Exploration lasts no more than $\epsilon T$ rounds. 
During exploitation, the algorithm identifies the estimated best arm among those searched and pulls it for the remainder of the game. 

\begin{algorithm}[ht]
\caption{$\epsilon$-Binary-Search-First}
\label{alg:epsbsf}
\textbf{Input:} $J$, $T$, R, $0 \leq \epsilon \leq 1$
\begin{algorithmic}[1]
\State Explore: Binary-Arm-Search($J$,$\lfloor \frac{\epsilon T}{\log J}\rfloor$,1,$J$,R)
\State Select arm $j$ such that $\hat r_j \geq R$ and $j \in \argmin_{i \in [J]} |\hat r_i - R|$
\For{remaining rounds $n \leq T$}
    \State Play arm $j$
\EndFor
\end{algorithmic}
\end{algorithm}

Binary-Arm-Search takes as input the number of arms $J$, the number of pulls $P$, a starting index, an ending index, and the threshold $R$. We set the number of pulls to be $P = \lfloor \frac{\epsilon T}{\log J}\rfloor.$

\begin{algorithm}[ht]
\caption{Binary-Arm-Search}
\label{alg:bas}

\textbf{Input:} $J$, $P$, $R$, Start, End

\begin{algorithmic}[1]
\If{End $\geq$ Start}
    \State Play arm $j = \lceil Start + \frac{End - Start}{2}\rceil$ for a total of $P$ times. Denote the empirical mean reward $\hat r_j.$
    \If{$\hat r_j \geq R$}
        \State Return Binary-Arm-Search($J$, $P$, Start, $j-1$, $R$)
    \Else
        \State Return Binary-Arm-Search($J$, $P$, $j+1$, End, $R$)
    \EndIf
\EndIf

\end{algorithmic}
\end{algorithm}

\subsection{What  Sequence: Opportunistic-TS}
\label{subsec:pur}
Consider a mobile user $t$. It is given a threshold $\hat{M}$ decided by the $\epsilon$-Binary-Search-First algorithm. Once the handover measurement procedure is triggered by $Y_{t,0} < \hat{M}$, we start to measure a sequence of target cells. Our goal is to determine the optimal sequence. 
While 
a classic multi-armed bandit algorithm can be used, we propose an opportunistic Thompson sampling algorithm motivated by a unique property observed in the real traces of extreme mobility. 

In particular, we observe that the change in signal strength over consecutive measurement is bounded (as empirically validated in Fig.\ref{fig:rsrp_dynamics} with the high-speed rails dataset).
This allows us to make the following \emph{regularity assumption} regarding the serving cell signal strength $Y_{t,n}$. We assume that there exists some positive constant $c$ such that $$|Y_{t,n} - Y_{t,n+1}| < c.$$ This ensures that the serving cell signal strength does not change ``too quickly" between consecutive measurements.  Under this assumption, we can have ``free'' measurements when the best target cell so far is good enough ($X_{best} \geq \hat{M}$) and the serving cell is still strong  enough ($Y_{t,n} \geq M + c$). So the next measurement is risk-free. Therefore, we can first find the best target cell and then use the ``free'' observations when available to satisfy the need for exploration.

The algorithm is outlined in Alg.~\ref{alg:pur}. This subroutine is executed for each mobile user $t$.
The algorithm takes as input the number of available neighboring cells $K$, the index of the mobile user $t$, and the handover threshold  $\hat{M}$ selected by $\epsilon$-Binary-Search-First.


\begin{algorithm}[ht]
\caption{Opportunistic Thompson Sampling (TS)} 
\label{alg:pur}
\textbf{Input:} $t,K, \hat M$, current TS posterior 

\textbf{Initialize:} $n = 0, X_{best} = 0, Y_{t,n} = \infty, B = \emptyset$, 
\begin{algorithmic}[1]
\If{$X_{best} < \hat M$}  
    \If{$Y_{t,n} > X_{best}$} 
        \State Measure target cell $I_{t,n}$ using TS, where   $I_{t,n} \not\in B$.
        \State Receive ($X_{I_{t,n}}$, $Y_{t,n}$)
        \State Update 
    \Else
        \State Handover to $X_{best}$
    \EndIf
\ElsIf{$Y_{t,n} \geq \hat{M}+c$}  \% ``free" observation
\State Measure target cell $I_{t,n}$ using TS  where $I_{t,n} \not\in B$.
        \State Receive ($X_{I_{t,n}}$, $Y_{t,n}$)
        \State Update 
\Else 
    \State Handover to $X_{best}$ 
\EndIf

\State def \textbf{Update}: 
\If{$X_{I_{t,n}} > X_{best}$}
    \State $X_{best} \leftarrow X_{I_{t,n}}$
\EndIf
    \State $n \leftarrow n+1$
    \State $B = B \cup I_{t,n}$
    \State Update TS posterior distribution of arm $I_{t,n}$

\end{algorithmic}
\end{algorithm}

This algorithm gives clear direction on how to make the handover vs. measurement decision. If the best target cell $X_{best}$ is not satisfactory (that is, $X_{best} < \hat M$) then the algorithm compares the serving cell to $X_{best}$ (Line 2). If $Y_{t,n} > X_{best}$, then the algorithm continues to measure  the best  unmeasured target selected using TS (Thompson Sampling). 
If $Y_{t,n} <X_{best}$, the mobile user $t$ handovers to $X_{best}$ (Line 7). 
Otherwise, as in line 9, if $X_{best} $ is satisfactory and $Y_{t,n} \geq \hat M + c,$ then the algorithm can make ``free" measurements. In this case, the algorithm explores by 
selecting an  unmeasured target selected using TS. We note that any bandit algorithms can be used in selecting target cells (to measure), such as UCB, greedy, and round-robin. In our experiments, we observe that TS achieves the best and most robust performance and thus adopt it here.

\section{Regret Analysis}
\label{sec:regret}

In this section, we analyze the regrets of \name's algorithms. 

\subsection{$\epsilon$-Binary-Search-First}
\label{subsec:CSAI}


Consider a $J$-armed stochastic bandit system. Recall from Sec.~\ref{sec:problem} that each arm $j$ has a random reward $f(Z_j) \in \{0,1\}$ with $\mathbb{E}[f(Z_j)] = r_j$ and $r_1 \leq r_2 \leq \cdots \leq r_J.$ The goal of the serving cell over $T$ rounds is to identify the closest sufficient arm $M$, as defined in Eq.~\ref{eq:M}.

Our problem requires a new definition of regret because, unlike the oracle, we have two types of losses that do not exist in the classic bandit: 1) a performance loss if we choose an arm that satisfies the constraint but is larger than necessary; and 2) a constraint violation. The traditional regret definition does not capture the fact that an optimal arm $a^*$ must be both the closest to and greater than the threshold value. To address this challenge, we define a new type of regret.

Let $N_T(j)$ be the number of times an arm $j$ is pulled under a given policy $\Gamma.$ We define the regret over $T$ rounds as $$R_\Gamma(T) = T - \mathbb{E}[N_T(a^*)] = \sum_{a \not= a^*} \mathbb{E}[N_T(a)].$$ 

Note that this definition of regret is ``coarser'' than the traditional stochastic bandit regret. This regret is an upper bound of the traditional regret, but is on the same order as it (that is to say, asymptotically, the two quantities will differ only by constant factor for the same policy). We need this coarser regret definition because it captures the performance of algorithms for closest sufficient arm identification in a manner that can be easily compared to results of algorithms for best arm identification.

We can now bound the regret accumulated by the $\epsilon$-Binary-Search-First algorithm. 
Define $\Delta = r_{M} - R$, $D = \min_{j} |r_M -r_j|$, and $d = \min_{j} |r_j -R|$, and $\delta = \min(\Delta, D/2)$. 

\begin{theorem}
Under the assumption of ordered arms, the $\epsilon$-Binary-Search-First strategy achieves regret bounded by\begin{align*}
    R(T) &\leq \log J \left(\frac{\log 6 \delta^2 T J}{2 \delta^2} - \frac{\log \log J}{2 \delta^2} + \frac{1}{2 \delta^2} + 1 \right)\\
\end{align*} under the assumption that $d < \sqrt{\frac{\log(T \log J)}{2 P}}$, where $d$ is the minimum absolute distance between a searched arm and $m$, and $\delta = \min(\Delta, D/2).$ 
\end{theorem}

 The proof is available in Appendix~\ref{subsec:epsregret}. Its intuition is as follows: we leverage the monotonicity assumption to reduce the number of arms searched. To address the noisy nature of the arm measurements, we pull each arm $\lfloor \epsilon T/\log{J} \rfloor$ times where $\epsilon$ is determined to minimize the overall regret, as shown in Appendix.
 We must show the following: first, the probability that an arm is not well-measured during binary search (defined as incorrectly identifying whether an arm's expected reward is larger or smaller than $R$) is small; and second, if all arms are well-measured during the binary search, then the best arm is explored. By combining these two parts and identifying an appropriate $\epsilon$, we achieve the above regret bound, as shown in detail in Appendix~\ref{subsec:epsregret}. 
 
\subsection{Opportunistic-TS}
\label{subsec:oppgreedy}

We note that the exact regret of the general ``what sequence" is difficult to evaluate, even if we use a classic UCB algorithm. The reason is that $Y_{t,n}$ is an unknown and non-stationary process over $n$.   However, a simpler case is where the mobile user is only allowed to measure one target cell and then handover to it. In this case, Opportunistic-TS reduces to the classic Thompson Sampling (TS) and thus yields $O(\log T)$ regret. We note that the ability to select among multiple target cells in general should yield better performance than classic TS in practice. This is also supported by our empirical evaluation using real traces.

\section{Experiments}
\label{sec:eval}

In this section, we evaluate \name's performance using real-world traces of extreme mobility and compare it with state-of-the-art handover policies and other bandit algorithms.

\begin{figure*}[t]
\vspace{0mm}
\subfloat[Reliability with $M$=-120 dBm]{
\includegraphics[width=.33\textwidth,valign=b]{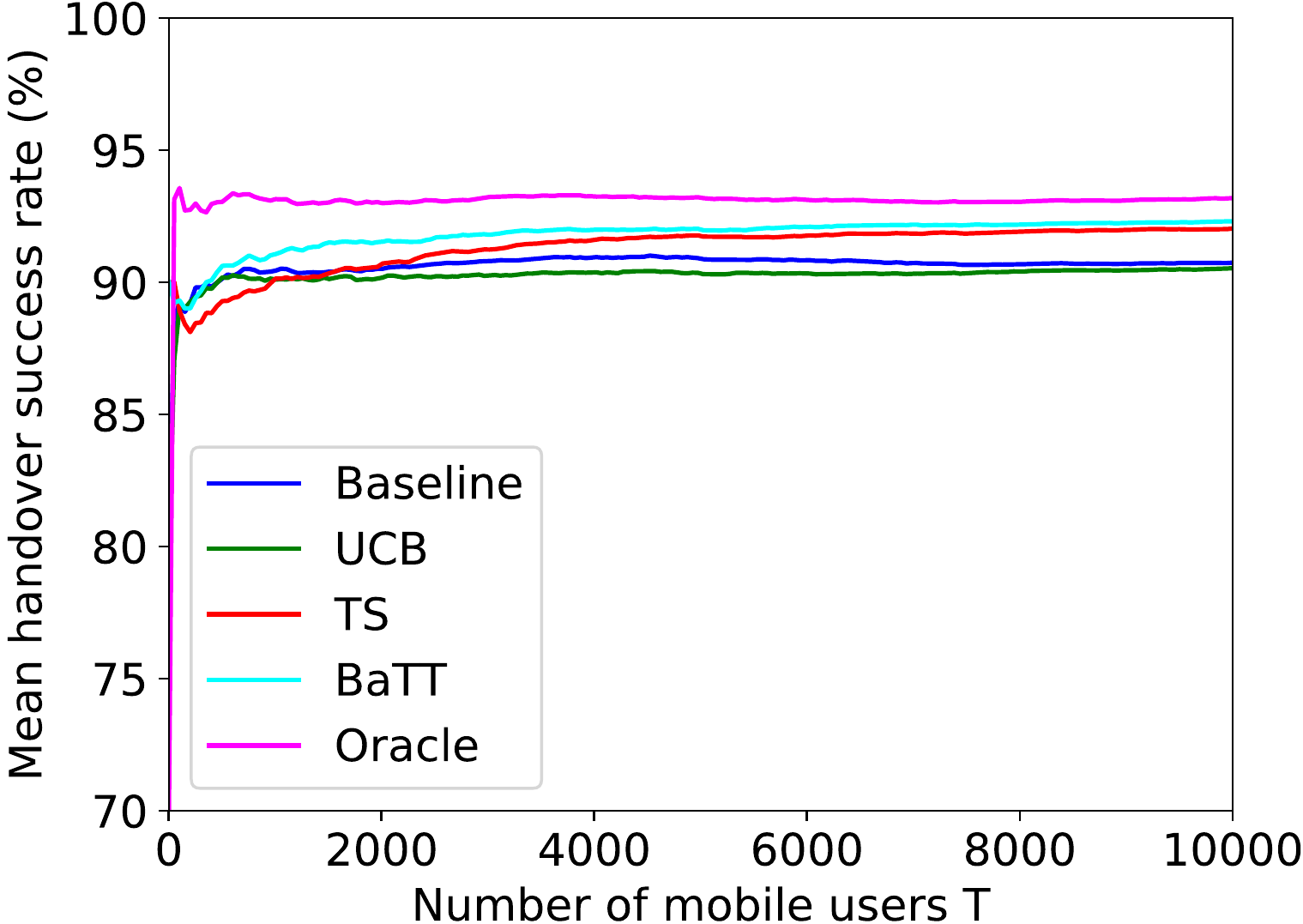}
\label{fig:eval:success_rate}
}
\subfloat[Regret with $M$=-120 dBm]{
\includegraphics[width=.33\textwidth,valign=b]{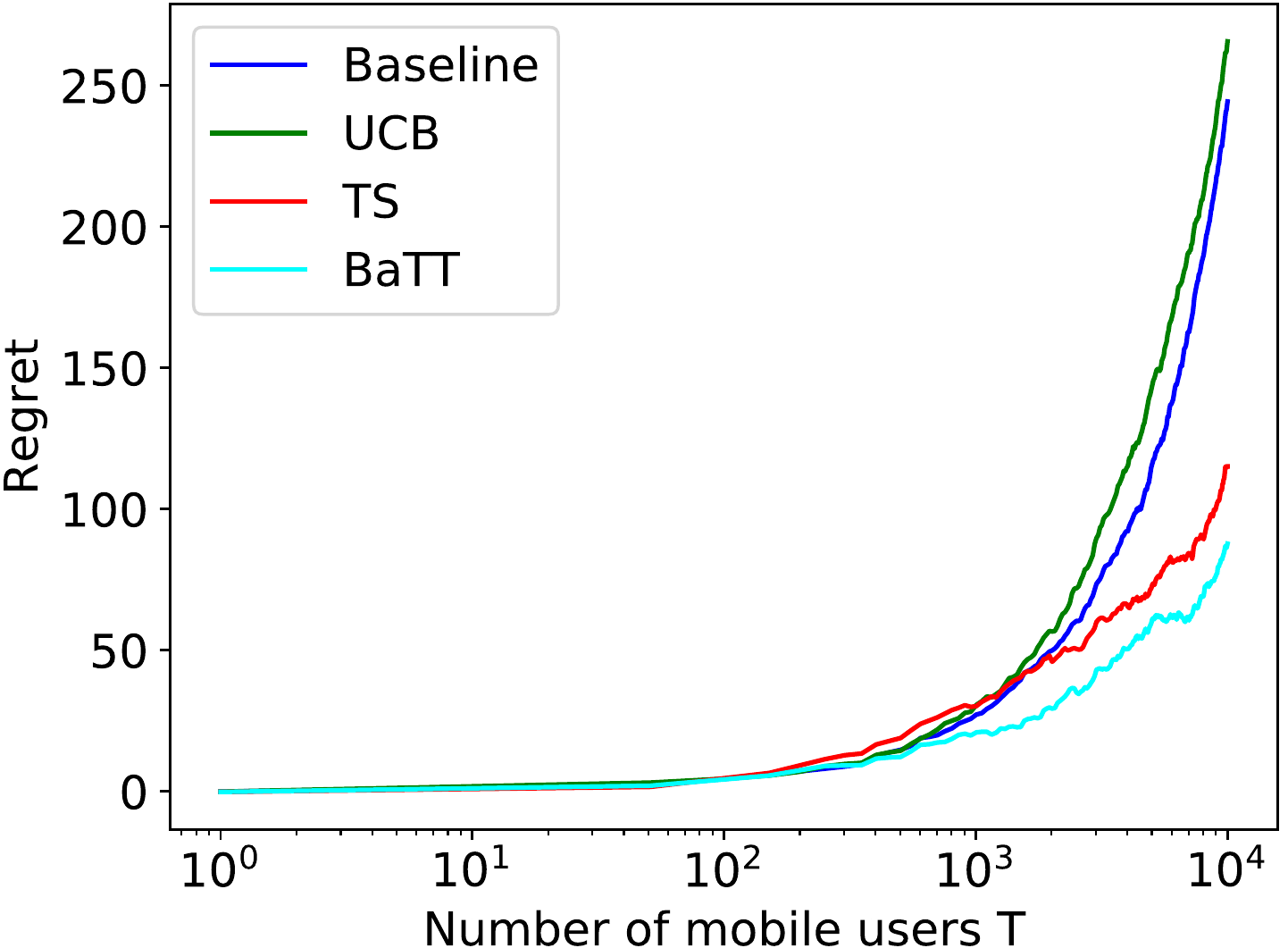}
\label{fig:eval:regret}
}
\subfloat[$\epsilon$-Binary-Search-First]{
\includegraphics[width=.33\textwidth,valign=b]{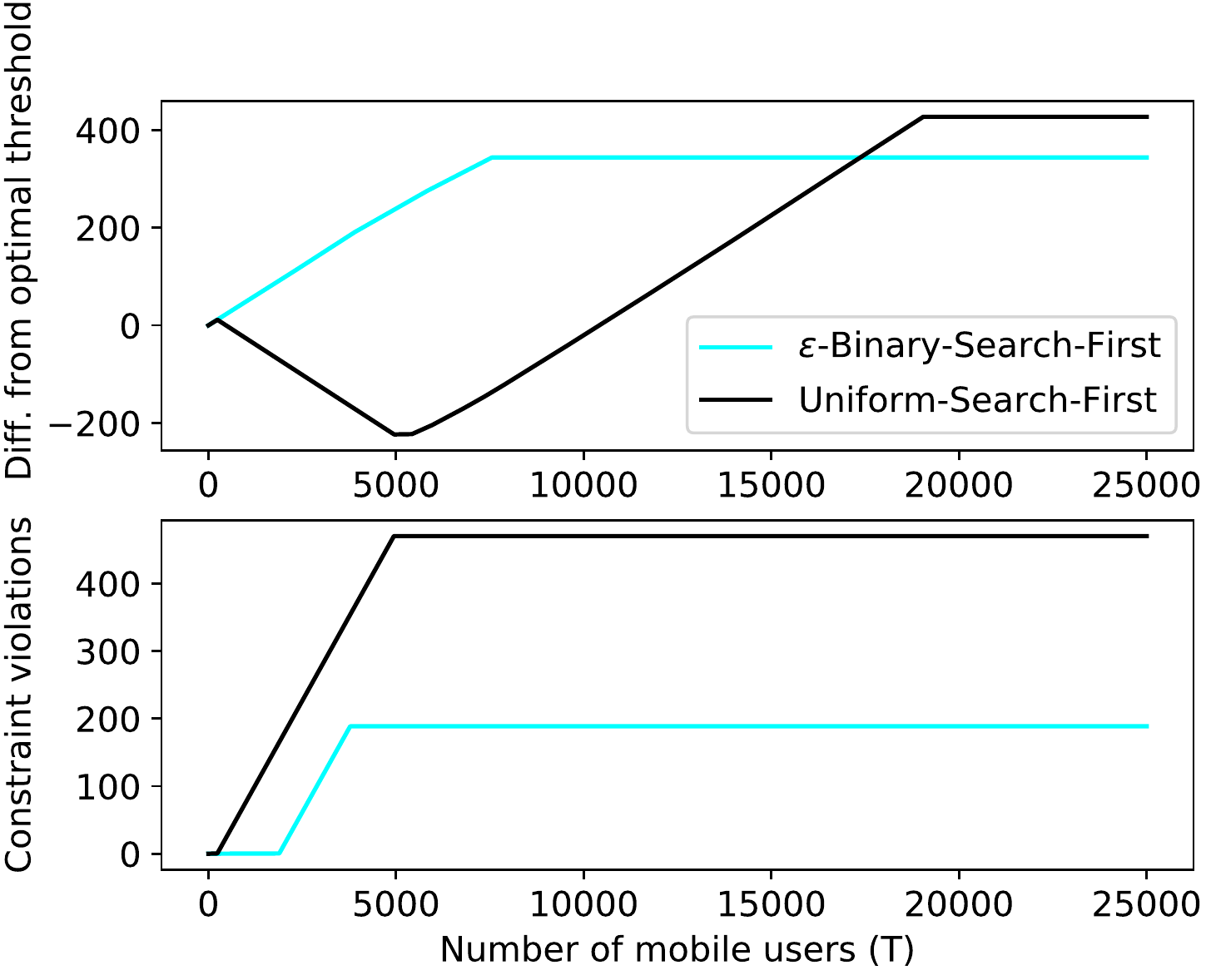}
\label{fig:eval:binary_search}
}
\caption{Comparison between \name, baseline LTE handover policy today, UCB, and Thompson sampling with $c$=4.}
\label{fig:eval}
\vspace{-5mm}
\end{figure*}

\subsection{Dataset}
\label{subsec:data}

We evaluate \name with a large-scale 4G LTE dataset on Chinese high-speed trains from \cite{wang2019active}. 
This dataset was collected on the high-speed rails between Beijing and Shanghai over 135,719 km of trips.
In these tests, a smartphone using China Mobile or China Telecom 4G LTE runs continuous {iperf} data transfer on the high-speed train at 200--350 km/h. 
Meanwhile, the smartphone also runs MobileInsight~\cite{li2016mobileinsight} to collect the 4G LTE signaling messages from the hardware cellular modem. 
These messages include 38,646 runtime measurement configurations of neighboring cell lists and thresholds, 81,575 measurement reports of serving/neighboring cell's signal strengths, and 23,779 handover commands from the serving cell to the mobile user device. As shown in  Figure~\ref{fig:validation}, these messages unveil real-world handover characteristics in extreme mobility and thus are helpful to evaluate \name.

\subsection{Regret and Benchmarks}
\label{subsec:regret}
We conduct a two-step evaluation of \name. First, we evaluate \name with a given threshold $M$. We then compare its regret and handover success rate with the following algorithms using the same threshold:

\begin{itemize}
    \item {\bf Oracle}: We assume that the average handover failure rates of the target cells are known. Therefore, we measure the target cells in an increasing order of the failure rate. 
    \item {\bf Baseline}: This is the state-of-the-art 4G/5G handover algorithm \cite{TS36.331,TS38.331,li2016mobileinsight}. It compares the serving cell and target cell's signal strengths and selects the first  neighboring cell with $X_{best}>Y_{t,n}$ as the target cell. This policy does not specify the ordering of cells to measure and relies on the device-specific cell scanning implementations instead \cite{TS36.304}. So, we assume the user's device measures the target cells randomly. 
    \item {\bf UCB}: We assume that the serving cell maintains UCB estimates for the target cells and instructs a mobile user to measure target cells based these estimates. 
    \item {\bf TS}: Similar to UCB, except using Thompson sampling. 
\end{itemize}
Note that in all these algorithms the handover happens when $X_{best}>Y_{t,n}$, that is, the best measured target cell is better than the current serving cell. The algorithms differ only in deciding the measurement order. 
The regret is defined as the handover failure rate difference between a the Oracle and a given algorithm. Recall that a failure happens when $f(Y_{t_n})=0$ or $g(X_{best})=0$, where we draw $f(\cdot)$ and $g(\cdot)$ from real traces as shown in Figure~\ref{fig:failure_vs_rsrp}. Because we use real traces, the evaluation results are not bounded by the assumptions made for the analysis. It thus provides a  realistic evaluation of the algorithms. 

Next, we assess \name's efficiency of threshold optimization. 
We compare \name's $\epsilon$-Binary-Search-First with a uniform threshold search.  

\subsection{Results}
\label{subsec:result}

We consider a serving cell with $K=10$ neighboring cells with the empirical signal strength distribution drawn from our dataset. Then we draw each cell's expected reward (handover success rate) by mapping their signal strength distribution to the handover success rate based on $f(\cdot)$ and $g(\cdot)$ from real traces. This results in the reward vector [0.76, 0.88, 0.90, 0.91, 0.92, 0.93, 0.94, 0.95, 0.97].
We then replay all sequences of serving cell's measurements before each handover command in the dataset.
For each serving cell's measurement, we run each algorithm to decide the next neighboring cell to measure. Then, we generate this neighboring cell's measurements based on its empirical distribution of signal strengths. With the serving/neighboring cell's measurements, each algorithm decides if the measurement should continue and selects the target cell if handover should start. 
Figure~\ref{fig:eval:success_rate},~\ref{fig:eval:regret} compare the handover success rates and regret. 

\subsubsection{Comparison with state-of-the-art}
\label{subsubsec:comp4g5g}

As shown in Figure~\ref{fig:eval:success_rate}, compared to the baseline in 4G/5G today, \name improves mean handover success rate from 89.7\% to 92.7\%. 
That is, it prevents 29.1\% handover failures in 4G/5G today. Further, note that \name approximates the Oracle, thus reaching the limit of the reliable mobility in 4G/5G today. Compared to the baseline, \name optimizes the ordering of the cells to measure when the serving cell's quality is decreasing, thus mitigating late handover failures. 

\subsubsection{Comparison with other bandit algorithms}
\label{subsubsec:compbandit}

Figure~\ref{fig:eval:success_rate} and \ref{fig:eval:regret} show \name outperforms UCB and TS. This is because \name adaptively balances the exploration and exploitation based on the runtime serving cell quality, while UCB and Thompson sampling do not. \name can accelerate the exploration when the serving cell quality is good and mitigate late handover failures when serving cell quality is not.
This is crucial, since late handovers due to the serving cells dominate the handover failures in reality, as shown in Table~\ref{tab:failure-ratio}. 

\subsubsection{Effectiveness of $\epsilon$-Binary-Search-First}

We consider a serving cell with $J = 81$ signal strength threshold values available to search from -140dBm to -60dBm. The empirical signal strength distribution is drawn from our dataset. We draw each the corresponding serving cell failure rate by mapping their signal strength distribution to the handover success rate based on $f(\cdot)$  from the real traces as shown in Figure~\ref{fig:failure_vs_rsrp}. 

We run $\epsilon$-Binary-Search-First over $T$ = 25000 rounds, with results averaged over $10$ trials. We compare $\epsilon$-Binary-Search-First to a uniform threshold tuning algorithm (Uniform-Search-First), which takes $\epsilon T$ rounds to uniformly sample all available arms and then picks the one with sample mean closest to optimal. 
We compare $\epsilon$-Binary-Search-First to Uniform-Search-First using two metrics: the cumulative number of constraint violations and the difference from the optimal threshold. 

Figure~\ref{fig:eval:binary_search} summarizes the results. 
It confirms $\epsilon$-Binary-Search-First outperforms Uniform-Search-First by having a smaller distance from the threshold overall, as well as plateauing sooner.
For 470 mobile users, Uniform-Search-First selects a serving cell signal strength that is lower than desirable. This is in contrast to the only 189 accumulated by $\epsilon$-Binary-Search-First. 
We also compute the cumulative difference from the optimal threshold value. We see that Uniform-Search-First accumulates a difference of -223.25 from the optimal threshold at T = 5000 before growing to 426.92. This is because Uniform-Search-First searches the arms in the order received. In contrast, $\epsilon$-Binary-Search-First accumulates only 343.62 cumulative difference from the optimal threshold. Moreover, $\epsilon$-Binary-Search-First stops accumulating error much sooner than Uniform-Search-First.

\section{Related Work}
\label{sec:related}

Reliable mobility management in 4G/5G cellular networks has been actively studied in recent years. Various deficiencies have been identified, such as sub-optimal radio coverage \cite{amaldi2008radio}, network misconfiguration \cite{deng2018mobility}, handover policy conflicts \cite{li2016instability, yuan2018resolving}, and late/blind handovers \cite{mismar2018partially}, to name a few. Our work studies a different aspect of handover failures in extreme mobility. In the context of extreme mobility, \cite{li2018measurement, wang2019active} report the non-negligible handover failures in reality and \cite{li2020beyond} unveils the exploration-exploitation tradeoff in handover decision policy and mitigates it by refining wireless communication paradigms. In contrast, our work moves further to explicitly address the exploration-exploitation trade-off using multi-armed bandits. 

There are some efforts to refine the performance of handover policy with machine learning techniques like XGBoost \cite{mismar2018partially}, fuzzy logic \cite{horrich2007adaptive}, neural networks \cite{horrich2007neural}, and support vector machines \cite{ong2010automated}. Our work differs from them because we focus on the handover reliability using multi-armed bandits. Bandit algorithms are well-suited to reliability in extreme mobility because they utilize online learning.

Our \name algorithm is motivated by recent advances in multi-armed bandits. Its problem formulation is inspired by the cascading bandit \cite{kveton2015cascading,zhou2018cost}, which returns a list of arms to pull.
The cost-aware cascading bandit considers a known cost of pulling each arm \cite{zhou2018cost} and is used for handover management \cite{wang2020neighbor}. Our problem differs from these because our evaluation is based on real-traces and does not assume a known cost.

Furthermore, our opportunistic-greedy algorithm is related to bandits with side observations. In \cite{degenne2018bandits}, a player can pay to observe before pulling an arm. In \cite{buccapatnam2017reward, pandey2007multi}, structured side information can be observed when pulling arm, e.g., on graphs. The focus is to leverage the structured information to decide which arm to pull. In comparison, we do not have such structures and our ``free'' observations depend on the environment. Specifically, our algorithm, motivated by opportunistic bandits \cite{wu2018adaptive}, leverages properties of the handover problem to obtain "free" observations. 

To the best of our knowledge, the constrained bandit problem formulation is new and so is our definition of the new regret that counts for both the performance loss of not choosing the best arm and the constraint violation. Our algorithm is a noisy version of binary search.  In the literature, algorithms have been developed for more sophisticated variations of the noisy observation problem \cite{karp2007noisy,nowak2009noisy}.
In comparison, our algorithm is simpler and with a formulation on constraint satisfaction that requires a different regret definition.

\section{Conclusion}
\label{sec:conc}

In this work, we strive for the reliable 4G/5G handover in extreme mobility using online learning techniques. 
We formulate the exploration-exploitation dilemma in extreme mobility as a multi-armed bandit problem. 
We propose the \name strategy to search for the optimal threshold of signal strength to address this dilemma {\em and} opportunistically balance the exploration and exploitation of target cells based on the runtime serving cell's signal strength. 
Our analysis shows $\epsilon$-Binary-Search-First achieves $\mathcal{O}(\log J \log T)$ regret, while opportunistic-TS achieves at most $\mathcal{O}(\log T)$ regret.
Experiments with real-world large-scale LTE datasets from the Chinese high-speed trains demonstrate \name's significant regret reduction and 29.1\% handover failure reduction. 

\bibliography{cellular}

\begin{thebibliography}{24}
\providecommand{\natexlab}[1]{#1}
\providecommand{\url}[1]{\texttt{#1}}
\providecommand{\urlprefix}{URL }
\expandafter\ifx\csname urlstyle\endcsname\relax
  \providecommand{\doi}[1]{doi:\discretionary{}{}{}#1}\else
  \providecommand{\doi}{doi:\discretionary{}{}{}\begingroup
  \urlstyle{rm}\Url}\fi

\bibitem[{3GPP(2019)}]{TS36.304}
3GPP. 2019.
\newblock {TS36.304: Evolved Universal Terrestrial Radio Access (E-UTRA); User
  Equipment (UE) procedures in idle mode}.

\bibitem[{3GPP(2020{\natexlab{a}})}]{TS36.331}
3GPP. 2020{\natexlab{a}}.
\newblock {TS36.331: Evolved Universal Terrestrial Radio Access (E-UTRA); Radio
  Resource Control (RRC)}.

\bibitem[{3GPP(2020{\natexlab{b}})}]{TS38.331}
3GPP. 2020{\natexlab{b}}.
\newblock {TS38.331: 5G NR: Radio Resource Control (RRC)}.

\bibitem[{Amaldi, Capone, and Malucelli(2008)}]{amaldi2008radio}
Amaldi, E.; Capone, A.; and Malucelli, F. 2008.
\newblock Radio planning and coverage optimization of 3G cellular networks.
\newblock \emph{Wireless Networks} 14(4): 435--447.

\bibitem[{Buccapatnam et~al.(2017)Buccapatnam, Liu, Eryilmaz, and
  Shroff}]{buccapatnam2017reward}
Buccapatnam, S.; Liu, F.; Eryilmaz, A.; and Shroff, N.~B. 2017.
\newblock Reward maximization under uncertainty: Leveraging side-observations
  on networks.
\newblock \emph{The Journal of Machine Learning Research} 18(1): 7947--7980.

\bibitem[{Degenne, Garcelon, and Perchet(2018)}]{degenne2018bandits}
Degenne, R.; Garcelon, E.; and Perchet, V. 2018.
\newblock Bandits with side observations: Bounded vs. logarithmic regret.
\newblock \emph{The Conference on Uncertainty in Artificial Intelligence (UAI)}
  .

\bibitem[{Deng et~al.(2018)Deng, Peng, Fida, Meng, and Hu}]{deng2018mobility}
Deng, H.; Peng, C.; Fida, A.; Meng, J.; and Hu, Y.~C. 2018.
\newblock Mobility Support in Cellular Networks: A Measurement Study on Its
  Configurations and Implications.
\newblock In \emph{Proceedings of the Internet Measurement Conference 2018},
  147--160. ACM.

\bibitem[{Horrich, Jamaa, and Godlewski(2007)}]{horrich2007adaptive}
Horrich, S.; Jamaa, S.~B.; and Godlewski, P. 2007.
\newblock Adaptive vertical mobility decision in heterogeneous networks.
\newblock In \emph{2007 Third International Conference on Wireless and Mobile
  Communications (ICWMC'07)}, 44--44. IEEE.

\bibitem[{Horrich, Jemaa, and Godlewski(2007)}]{horrich2007neural}
Horrich, S.; Jemaa, S.~B.; and Godlewski, P. 2007.
\newblock Neural networks for adaptive vertical handover decision.
\newblock In \emph{2007 5th International Symposium on Modeling and
  Optimization in Mobile, Ad Hoc and Wireless Networks and Workshops}, 1--7.
  IEEE.

\bibitem[{Karp and Kleinberg(2007)}]{karp2007noisy}
Karp, R.~M.; and Kleinberg, R. 2007.
\newblock Noisy binary search and its applications.
\newblock In \emph{Proceedings of the eighteenth annual ACM-SIAM symposium on
  Discrete algorithms}, 881--890.

\bibitem[{Kveton et~al.(2015)Kveton, Szepesvari, Wen, and
  Ashkan}]{kveton2015cascading}
Kveton, B.; Szepesvari, C.; Wen, Z.; and Ashkan, A. 2015.
\newblock Cascading bandits: Learning to rank in the cascade model.
\newblock In \emph{International Conference on Machine Learning}, 767--776.

\bibitem[{Li et~al.(2018)Li, Xu, Li, Zheng, Peng, Wang, Wang, Shen, and
  Mijumbi}]{li2018measurement}
Li, L.; Xu, K.; Li, T.; Zheng, K.; Peng, C.; Wang, D.; Wang, X.; Shen, M.; and
  Mijumbi, R. 2018.
\newblock A measurement study on multi-path tcp with multiple cellular carriers
  on high speed rails.
\newblock In \emph{Proceedings of the 2018 Conference of the ACM Special
  Interest Group on Data Communication}, 161--175.

\bibitem[{Li et~al.(2016{\natexlab{a}})Li, Deng, Li, Peng, and
  Lu}]{li2016instability}
Li, Y.; Deng, H.; Li, J.; Peng, C.; and Lu, S. 2016{\natexlab{a}}.
\newblock Instability in distributed mobility management: Revisiting
  configuration management in 3g/4g mobile networks.
\newblock In \emph{ACM SIGMETRICS Performance Evaluation Review}, volume~44,
  261--272. ACM.

\bibitem[{Li et~al.(2020)Li, Li, Zhang, Baig, Qiu, and Lu}]{li2020beyond}
Li, Y.; Li, Q.; Zhang, Z.; Baig, G.; Qiu, L.; and Lu, S. 2020.
\newblock Beyond 5G: Reliable Extreme Mobility Management.
\newblock In \emph{Proceedings of the ACM Special Interest Group on Data
  Communication (SIGCOMM)}, 344--358. ACM.

\bibitem[{Li et~al.(2016{\natexlab{b}})Li, Peng, Yuan, Li, Deng, and
  Wang}]{li2016mobileinsight}
Li, Y.; Peng, C.; Yuan, Z.; Li, J.; Deng, H.; and Wang, T. 2016{\natexlab{b}}.
\newblock Mobileinsight: Extracting and Analyzing Cellular Network Information
  on Smartphones.
\newblock In \emph{Proceedings of the 22nd Annual International Conference on
  Mobile Computing and Networking}, MobiCom '16. ACM.

\bibitem[{Mismar and Evans(2018)}]{mismar2018partially}
Mismar, F.~B.; and Evans, B.~L. 2018.
\newblock Partially blind handovers for mmWave new radio aided by sub-6 GHz LTE
  signaling.
\newblock In \emph{2018 IEEE International Conference on Communications
  Workshops (ICC Workshops)}, 1--5. IEEE.

\bibitem[{Nowak(2009)}]{nowak2009noisy}
Nowak, R. 2009.
\newblock Noisy generalized binary search.
\newblock In \emph{Advances in neural information processing systems},
  1366--1374.

\bibitem[{Ong, Magrabi, and Coiera(2010)}]{ong2010automated}
Ong, M.-S.; Magrabi, F.; and Coiera, E. 2010.
\newblock Automated categorisation of clinical incident reports using
  statistical text classification.
\newblock \emph{Quality and Safety in Health Care} 19(6): e55--e55.

\bibitem[{Pandey, Chakrabarti, and Agarwal(2007)}]{pandey2007multi}
Pandey, S.; Chakrabarti, D.; and Agarwal, D. 2007.
\newblock Multi-armed bandit problems with dependent arms.
\newblock In \emph{Proceedings of the 24th international conference on Machine
  learning}, 721--728.

\bibitem[{Wang et~al.(2020)Wang, Yang, He, Zhou, Chen, and
  Jiang}]{wang2020neighbor}
Wang, C.; Yang, J.; He, H.; Zhou, R.; Chen, S.; and Jiang, X. 2020.
\newblock Neighbor Cell List Optimization in Handover Management Using
  Cascading Bandits Algorithm.
\newblock \emph{IEEE Access} 8: 134137--134150.

\bibitem[{Wang et~al.(2019)Wang, Zheng, Ni, Xu, Qian, Li, Jiang, Cheng, Cheng,
  Li et~al.}]{wang2019active}
Wang, J.; Zheng, Y.; Ni, Y.; Xu, C.; Qian, F.; Li, W.; Jiang, W.; Cheng, Y.;
  Cheng, Z.; Li, Y.; et~al. 2019.
\newblock An Active-Passive Measurement Study of TCP Performance over LTE on
  High-speed Rails.
\newblock In \emph{ACM MobiCom}.

\bibitem[{Wu, Guo, and Liu(2018)}]{wu2018adaptive}
Wu, H.; Guo, X.; and Liu, X. 2018.
\newblock Adaptive exploration-exploitation tradeoff for opportunistic bandits.
\newblock In \emph{International Conference on Machine Learning}, 5306--5314.

\bibitem[{Yuan et~al.(2018)Yuan, Li, Li, Lu, Peng, and
  Varghese}]{yuan2018resolving}
Yuan, Z.; Li, Q.; Li, Y.; Lu, S.; Peng, C.; and Varghese, G. 2018.
\newblock Resolving Policy Conflicts in Multi-Carrier Cellular Access.
\newblock \emph{cell} 2(C1): 4G.

\bibitem[{Zhou et~al.(2018)Zhou, Gan, Yan, and Shen}]{zhou2018cost}
Zhou, R.; Gan, C.; Yan, J.; and Shen, C. 2018.
\newblock Cost-aware cascading bandits.
\newblock \emph{arXiv preprint arXiv:1805.08638} .

\end{thebibliography}
\bibliographystyle{aaai}

\newpage

\newpage
\begin{appendix}
\label{sec:appendix}

\section{Proof of Theorem 5.1} 
\label{subsec:epsregret}

To prove Theorem 5.1, we first prove the lemma below. An arm $a$ is well-measured if $r_a \geq R$ and $\hat r_a \geq R$ or if $r_a < R$ and $\hat r_a < R.$
\begin{lemma}
If all arms are well-measured, then the best arm will be measured. 
\end{lemma}
\label{subsec:lemma}
\begin{proof}
We prove the statement by contradiction. Suppose that the best arm $a^*$ is not measured during exploration. Then, the final interval between Start and End points considered by Binary-Arm-Search must be either completed above or below the arm $a^*$. That is to say, both Start and End must be to the right of $a^*$ or both to the left of $a^*$. Without loss of generality, suppose that both Start and End are to the right $a^*$. It must be the case that the sample means $r_{Start} \geq 0$ and $r_{End} \geq 0.$ This is a contradiction, since this means either Start or End is not well-measured. 
\end{proof}

We next prove Theorem 5.1. We first decompose regret into regret accumulated during exploration and exploitation: \begin{align*}
    R(T) &= \sum_{a\not= a^*} \mathbb{E}[N_T(a)]\\
    &\leq \epsilon T + (1 - \epsilon) T \sum_{a \not= a^*} \mathbb{P}[I = a],
\end{align*} where $I$ indicates the arm chosen for exploitation.

It is necessary to bound the term $\mathbb{P}[I = a]$ for each $a$, that is, the probability that the algorithm chooses a suboptimal arm $a$ for exploitation. Let $a^*$ be the arm such that $p_{a^*} - R \geq 0$ and $a^* \in \argmin_{k \in [K]} p_k - R$. Our analysis consists of two parts: computing regret with the assumption that $a^* \in S$ (that is, the optimal arm is searched during exploration) and showing that the probability that the optimal arm is not searched during exploration, $\mathbb{P}[a^* \not\in S]$, is negligible. 

For the sake of the analysis, we introduce the following notation.  Let $S$ be the set of arms searched during the exploration phase. Clearly $S \subset [J]$ and $|S| \leq \lceil \log J \rceil$. Each arm $j \in S$ has been searched $P = \lfloor \frac{\epsilon T}{\log J}\rfloor$ times. Let $A$ be the set of arms $a$ such that $r_a \geq R$. Let $B$ be the set of arms $b$ such that $r_b < R$. Let $\Delta_a = r_a - R > 0$. Let $\Delta_a' = R - r_a > 0$. For all arms $j \in S$, let $D_j = |r_j - r_{j^*}|$. 

For the first part of the proof, we assume that the optimal arm is searched. That is, $a^* \in S$. Now, fix some suboptimal arm $a \in S$. Let $\hat r_a$ denote the sample mean of arm $a$ after exploration and let $r_a$ denote the true mean of arm $a$. Finally, $R$ is the handover threshold. Then, for $a$ to be chosen for exploitation, one of two cases must hold: \begin{enumerate}
    \item The sample mean of the optimal arm must be underestimated, $\hat r_{a^*} < R$, and the sample mean of the suboptimal arm must be greater than the threshold value, $R < \hat r_a$; or 
    \item the original arm is well-estimated, but the sample mean of the suboptimal arm is smaller than that of the optimal arm, that is, $R < \hat r_a < \hat r_{a^*}.$
\end{enumerate}

Therefore, for a fixed suboptimal arm $a \in S$,\begin{align*}
    \mathbb{P}[I = a] &\leq \mathbb{P}[\hat r_{a^*} < R < \hat r_a] + \mathbb{P}[R < \hat r_a < \hat r_{a^*}] \\
           &\leq \mathbb{P}[\hat r_{a^*} < R] \mathbb{P}[R < \hat r_a] + \mathbb{P} [ \hat r_a < \hat r_{a^*}] 
\end{align*} By Hoeffding's inequality, $$\mathbb{P}[\hat r_{a^*} < R] = \mathbb{P}[\hat r_{a^*} \leq r_{a^*} - \Delta] \leq e^{-2 P \Delta^2}.$$ Trivially, it is true that $\mathbb{P}[R < \hat r_a] \leq 1.$

Finally, consider the event $[\hat r_a < \hat r_{a^*}].$ We claim that the event $[\hat r_a < \hat r_{a^*}]$ is a subset of the event $[\hat r_{a^*} \geq r_{a^*} + \frac{D_a}{2}] \cup [\hat r_a < r_a - \frac{D_a}{2}].$  To see this, assume for the sake of contradiction that the complement $[[\hat r_{a^*} \geq r_{a^*} + \frac{D_a}{2}]\cup [\hat r_a < r_a - \frac{D_a}{2}]]^\complement$ is true. Then, $\hat r_{a^*} < r_{a^*} + \frac{D_a}{2}$ and $\hat r_a  > r_a - \frac{D_a}{2}.$ It follows that $\hat r_{a^*} < r_a + \frac{D_a}{2}$ and $\hat r_a > r_{a^*} - \frac{D_a}{2}$. Now observe that $r_a + \frac{D_a}{2} = r_{a^*} - \frac{D_a}{2}$. Thus, $\hat r_{a^*} < r_{a^*} - \frac{D_a}{2} < \hat r_a$. Thus we have proven the claim. Therefore, by the inclusion law, it is the case that $\mathbb{P}[\hat r_a < \hat r_{a^*}] \leq \mathbb{P}[[\hat r_{a^*} > r_{a^*} + \frac{D_a}{2} ]\cup [\hat r_a < r_a - \frac{D_a}{2}]]$. Finally, the union bound implies that \begin{align*}
    \mathbb{P}[\hat r_a < \hat r_{a^*}] &\leq \mathbb{P}[\hat r_{a^*} > r_{a^*} + \frac{D_a}{2}] + \mathbb{P}[\hat r_a < r_a - \frac{D_a}{2}] \\
    &\leq \mathbb{P}[\hat r_{a^*} > r_{a^*} + \frac{D}{2}] + \mathbb{P}[\hat r_a < r_a - \frac{D}{2}] \\
    &\leq 2 e^{-2 P (D/2)^2}
 \end{align*} with the final step following from Hoeffding's inequality. Thus, $$\mathbb{P}[I = a] \leq e^{-2 P \Delta^2} + 2 e^{-2 N (D/2)^2}.$$ Recall that we define $\delta = \min\{D/2, \Delta\}.$Then, $$\mathbb{P}[I = a] \leq e^{-2 N \Delta^2} + 2 e^{-2 P (D/2)^2} \leq 3 e^{-2 P \delta^2}.$$  We can therefore express the regret bound as $$R(T) \leq \epsilon T + 3TJ e^{-2 P \delta^2} \leq \epsilon T + 3TJ e^{-2 (\frac{\epsilon T}{\log J} -1) \delta^2}.$$
 
 We wish to find the optimal value of $\epsilon$ to minimize the regret bound found above. Let $$h(\epsilon) = \epsilon T + 3TJ e^{-2 (\frac{\epsilon T}{\log J} -1) \delta^2}.$$ Then its derivative with respect to $\epsilon$ is  $$h'(\epsilon) = T + 3TJ \left( \frac{-2 \delta^2 T}{\log J}\right)e^{-2 \delta^2 \left(  \frac{\epsilon T}{\log J} - 1\right)}.$$ Setting $h'(\epsilon)$ equal to $0$ and solving for $\epsilon$, we arrive at the optimal value $$\epsilon = \frac{\log J}{T} - \frac{\log J}{2 T \delta^2} \log\left( \frac{\log J}{6 \delta^2 TJ} \right).$$ Substituting this value of $\epsilon$ into the regret upper bound $h(\epsilon)$, we see that $$R(T) \leq \log J \left(\frac{\log 6 \delta^2 TJ}{2 \delta^2} - \frac{\log \log J}{2 \delta^2} + \frac{1}{2 \delta^2} + 1 \right).$$ Observe that the regret is on the order of $\mathcal{O}\left((\log J)(\log T) \right).$
 
Having established the upper bound on regret with the assumption that the optimal arm $a^*$ is searched, we now bound the probability that $a^*$ is missed during the exploration phase. That is, we compute and bound $\mathbb{P}[a^* \not\in S]$. Per the lemma above, if $a^* \not\in S$ then some arm $a \in S$ must not be well-measured. That is, binary search estimated some arm $a$ to have $\hat r_a < R$ when in fact, $r_a \geq R$ or vice versa. Let $S_a$ be the event that arm $a$ is searched and well-measured. Then, $$\mathbb{P} [a^* \not\in S] \leq 1 - \mathbb{P} [ \cap_{a \in S} S_a].$$ 

Consider some $a \in S \cap A$. Since $r_a - R \geq 0$, by Hoeffding's inequality, $$\mathbb{P}[\hat r_a \leq R] = \mathbb{P}[\hat r_a \leq r_a - \Delta_a] \leq e^{-2 P \Delta_a^2}.$$ Therefore, for $a \in S \cap A$, $\mathbb{P}[S_a] \geq 1 - e^{-2 P \Delta_a^2}.$ Similarly, for $b \in B \cap S$, we know $R - r_b \geq 0$. By Hoeffding's inequality, $$\mathbb{P}[\hat r_b \geq R] = \mathbb{P}[\hat p_b \geq p_b + \Delta_b'] \leq e^{-2 P \Delta_b'^2}.$$ Thus, for $b \in S \cap B$, $\mathbb{P}[S_b] \geq 1 - e^{-2 P \Delta_b'^2}$. By the reverse union bound, $$\mathbb{P}[\cap_{a \in S} S_a] \geq 1 - \sum_{a \in S\cap A} e^{-2 P \Delta_a^2} - \sum_{b \in S \cap B} e^{-2 P \Delta_b'^2}.$$ We conclude that \begin{align*}
    \mathbb{P}[a^* \not\in S] &\leq 1 - \left( 1 - \sum_{a \in S\cap A} e^{-2 P \Delta_a^2} - \sum_{b \in S \cap B} e^{-2 P \Delta_b'^2} \right)\\
    &= \sum_{a \in S\cap A} e^{-2 P \Delta_a^2} + \sum_{b \in S \cap B} e^{-2 P \Delta_b'^2}.
\end{align*} 

Recall that we define $d = \min_{a, b} \{\Delta_a, \Delta_b'\}.$ Then, $$\mathbb{P}[a^* \not\in S] \leq \log J e^{-2 P d^2},$$ which is negligible for $d < \sqrt{\frac{\log(T \log J)}{2P}}.$ 

\end{appendix}

\end{document}